# Performance Evaluation of Different Techniques for texture Classification

Ashwini Dange[1], Mugdha Khade[2], Payal Kulkarni[3], Pooja Maknikar[4]

[1]Department of Electronics Engineering, VIT, Pune
ashwini.dange8@gmail.com
mugdha_797@yahoo.co.in
payalkulkarni31@yahoo.co.in
pooja_maknikar@yahoo.in

**ABSTRACT**

*Texture is the term used to characterize the surface of a given object or phenomenon and is an important feature used in image processing and pattern recognition. Our aim is to compare various Texture analyzing methods and compare the results based on time complexity and accuracy of classification. The project describes texture classification using Wavelet Transform and Co occurrence Matrix. Comparison of features of a sample texture with database of different textures is performed. In wavelet transform we use the Haar, Symlets and Daubechies wavelets. We find that, thee 'Haar' wavelet proves to be the most efficient method in terms of performance assessment parameters mentioned above. Comparison of Haar wavelet and Co-occurrence matrix method of classification also goes in the favor of Haar. Though the time requirement is high in the later method, it gives excellent results for classification accuracy except if the image is rotated.*

## 1. INTRODUCTION

[4]Texture is that innate property of all surfaces that describes visual patterns, and that contain important information about the structural arrangement of the surface and its relationship to the surrounding environment. Texture consists of primitive or texture elements called as texels.

Texture analysis is important in many applications of computer image analysis for classification or segmentation of images based on local spatial variations of intensity. The important task in texture classification is to extract texture features which most completely embody the information of texture in the original image. Here we study two texture classification methods: Wavelet Transform and Co-occurrence matrix.

### 1.1 Wavelet Transform

Main feature of wavelet is Multi-resolution analysis and variable window size. Narrow window gives good time and poor frequency resolution ,wide window gives good frequency resolution and bad time resolution.

The WT has a good time and poor frequency resolution at high frequencies, and good frequency and poor time resolution at low frequencies. [1]

## 1.2 Co-occurrence matrix

A co-occurrence matrix C is an n×n matrix, where n is the number of gray-levels within the image. The matrix C (i , j) counts the number of pixel pairs having the intensities i and j. These pixel pairs are defined by specified distances and directions, which can be represented by a displacement vector d = (dx, dy), representing the number of pixels between the pair in the x and y directions.

## 2. WAVELET

One method of representing image is in the form of 2 dimensional array, each value representing brightness of the pixel. Smooth variations in these values are termed as Low freq components and sharp variations as High freq. components. Low freq components constitute base of an image and High frequency components refine them. In short smooth variations demand more importance than the details.

A fundamental goal of data compression is to reduce the bit rate for transmission or storage while maintaining an acceptable fidelity or image quality. Compression can be achieved by transforming the data, projecting it on the basis of functions, and then encoding this transform.

The basic idea of wavelet transform is to represent any arbitrary function $f$ as a superposition of wavelets. Any such superposition decomposes $f$ into different scale levels, where each level is then further decomposed with a resolution adapted to the level. One way to achieve such a decomposition writes $f$ as an integral over a and b with appropriate weighting coefficients

$$\psi^{(a,b)} * t = mod\ (a)^{-0.5} * \psi \frac{(t-b)}{a}$$

where,

a=scale (1/frequency)

b=time shift.

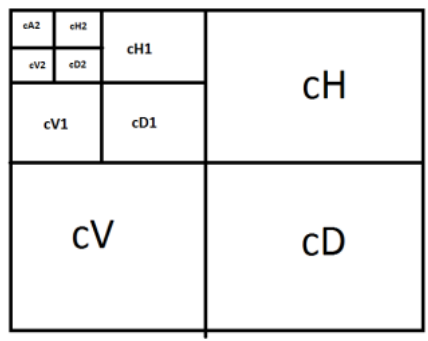

Figure 1

By decomposing the image into a series of high-pass and low pass bands, the wavelet transform extracts directional details that capture horizontal (cH), vertical (cV) and the diagonal (cD) activity. Since lower spatial frequencies of an image are more significant for the image's characteristics than higher spatial frequencies, further filtering of the approximation is useful.

Types of wavelet

## 2.1 Haar Wavelet Transform

[2]A Haar wavelet is the simplest type of wavelet. In discrete form, Haar wavelets are related to a mathematical operation called the Haar transform. The Haar transform serves as a prototype for all other wavelet transforms. Like all wavelet transforms, the Haar transform decomposes a discrete signal into two sub signals of half its length. One sub signal is a running average or trend; the other sub signal is a running difference or fluctuation. The Haar wavelet transform has a number of advantages such as it is conceptually simple, fast, memory efficient.

[7] The Haar transform is based on the Haar functions, $h_k(z)$, which are defined over the continuous, closed interval [0,1] for z, and for k=0,1,2,…,N-1, where $N=2^n$. The first step in generating the Haar transform is to note that the integer *k* can be decomposed uniquely as

$k=2^p+q-1$

where 0≤p≤n-1, and

q=0 or 1 for p=0, and $1 \le q \le 2^p$ for p≠0.

With this background, the Haar functions are defined as

$$h_0 \triangleq h_{00}(z) = \frac{1}{\sqrt{N}} \qquad \text{for } z \in [0,1]$$

and

$$h_k(z) \triangleq h_{00}(z) = \frac{1}{\sqrt{N}} \begin{cases} 2^{p/2} & \frac{q-1}{2^p} \le z \propto \frac{q-1/2}{2^p} \\ -2^{p/2} & \frac{q-1/2}{2^p} \le z \propto \frac{q}{2^p} \\ 0 & \text{otherwise for } z \in [0,1] \end{cases}$$

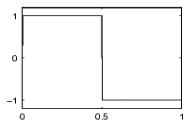

Haar wavelet is discontinuous, and resembles a step function. The Haar transform also has limitations, which can pose a problem for some applications. In generating each of the averages for the next level and each set of coefficients, the Haar transform performs an average and difference on a pair of values. Then the algorithm shifts over by two values and calculates another average and difference on the next pair. The high frequency coefficient spectrum should reflect all high frequency changes.

## 2.2  Daubechies Wavelet Transform

[2]The Daubechies wavelet transforms are defined in the same way as the Haar wavelet transform by computing the running averages and differences via scalar products with scaling signals and wavelets, the only difference between them consists in how these scaling signals and wavelets are defined.

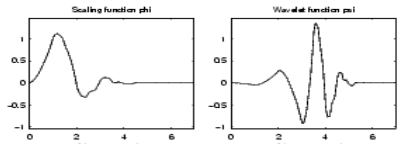

This wavelet type has balanced frequency responses but non-linear phase responses. Daubechies wavelets use overlapping windows, so the high frequency coefficient spectrum reflects all high frequency changes. Therefore Daubechies wavelets are useful in compression and noise removal of audio signal processing.

### 2.3 Sym8

Symlets are generally written as symN where N is the order. Some authors use 2N instead of N. Symlets are only near symmetric, we used sym8 in this experiment.

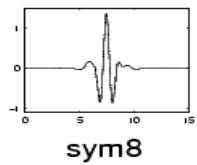

General characteristics: Compactly supported wavelets with least asymmetry and highest number of vanishing moments for a given support width. Associated scaling filters are near linear-phase filters.

### 3. CO-OCCURRENCE MATRIX

Gray level co-occurrence matrix (GLCM) has been proven to be a very powerful tool for texture image segmentation. The only shortcoming of the GLCM is its computational cost [8]. GLCM is a matrix that describes the frequency of one gray level appearing in a specified spatial linear relationship with another gray level within the area of investigation [8].

The following figure shows how gray co-matrix calculates several values in the GLCM of the 4-by-5 image I. Element (1,1) in the GLCM contains the value 1 because there is only one instance in the image where two, horizontally adjacent pixels have the values 1 and 1. Element (1,2) in the GLCM contains the value 2 because there are two instances in the image where two, horizontally adjacent pixels have the values 1 and 2. Gray co-matrix continues this processing to fill in all the values in the GLCM[9].

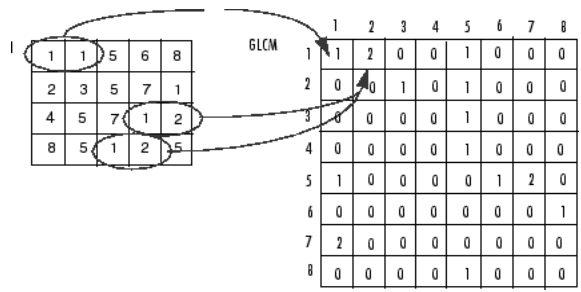

Mathematically, a co-occurrence matrix C is defined over an n x m image I, parameterized by an offset (Δx,Δy),[10]

$$C_{\Delta x, \Delta y}(i,j) = \sum_{p=1}^{n}\sum_{q=1}^{m} \begin{cases} 1, & \text{if } I(p,q) = i \text{ and if } I(p+\Delta x, q+\Delta y) = j \\ 0, & \text{otherwise} \end{cases}$$

# 4. METHODOLOGY

The texture classification algorithm consists of three main steps:

1. Segmentation of regions of interest,

2. Extraction of the most discriminative texture features, and

3. Creation of a classifier that automatically identifies the various textures.

## 4.1 Algorithm

### 4.1.1 for Wavelet

1. Standard images are collected and a database is formed for classification.
2. The number of decomposition levels was taken to be 3 and the steps 3, 4 and 5 were carried out for level times i.e. 3 times.
3. Wavelet transform is applied to each of the image in the database. Each of the types such as Haar, Sym8 and DB4 are applied and features of the image are extracted.
4. The energyis calculated by dividing addition of all elements of the array CH/CV by the number of rows multiplied by number of columns in the CH/CV array.
5. If number of levels is reached then calculate energy for CA following the steps 3 and 4.
6. The calculated energy values are stored in an array called featureset. This array will have seven columns as it is a three level decomposition and the number of rows is equivalent to the number of images.
7. An image that is to be classified is considered; noise is added, contrast is changed by histogram equalization. These three images are taken as inputs for the classification.
8. The features of these images are extracted and stored in an array called new feature set following the steps 3, 4 and 5 above.
9. This new feature set array is compared with the database array, feature set using the concept of Euclidean distance. This is done by using the function 'dist'.
10. The minimum value of the distances is calculated and according to the threshold value the classes are determined.

### 4.1.2 For Co-occurrence

1. Standard images are collected and a database is formed for classification.
2. We form the co-occurrence matrix of the images in the database along 0 degrees, 45 degrees, 90 degrees, 135 degrees along both the positive and negative axis.
3. The energy of these four matrices is calculated by addition of square of each element in the four matrices.
4. The calculated energy values are stored in an array, say, feature set. This array will have four columns as it is formed from four matrices.
5. An image that is to be classified is considered; noise is added, contrast is changed by histogram equalization. The images are also rotated at various angles. These images are taken as inputs for the classification.
6. The features of these images are extracted and stored in an array, say newfeatureset, following the steps 2, 3, 4 and 5 above.

7. This new feature set array is compared with the database array, featureset using the concept of Euclidean distance. This is done by using the function 'dist'.
8. The minimum value of the distances is calculated and according to the threshold value the classes are determined.

## 5. RESULT TABLES

### 5.1 Average time for Computations

#### 5.1.1 Original Image:

Table 1

| Transform | Db4 | Sym8 | Haar | Co-occurence |
|---|---|---|---|---|
| Time (Sec) | 0.1986 | 0.221 | 0.1805 | 4.2799 |

### 5.1.2 Noisy Images (Salt & Pepper):

Table 2

| Noise(Salt and paper) | Haar | Co-occurrence |
|---|---|---|
| 0.02 | 0.1844 | 4.344 |
| 0.05 | 0.183 | 4.344 |
| 0.09 | 0.2097 | 4.3305 |

#### 5.1.3 Image with changed contrast (Histogram Equalised):

Table 3

| Transform | Haar | Co-occurrence |
|---|---|---|
| Time(sec) | 0.2077 | 4.39 |

#### 5.1.4 Rotated Image

Table 4

| Rotation | Haar | Co-occurrence |
|---|---|---|
| 2 degree | 0.1914 | 0.713 |
| 4 degree | 0.1915 | 0.73 |
| 30 degree | 0.193 | 0.7346 |

### 5.2 Accuracy

#### 5.2.1 Original Image

Table 5

| Db4 | Sym8 | Haar | Co-occurrence |
|---|---|---|---|
| 100% | 100% | 100% | 100% |

**5.2.2 Noisy Images**

Table 6

| Noise (Salt and pepper) | Haar | Co-occurrence |
|---|---|---|
| 0.02 | 78% | 70% |
| 0.05 | 70% | 22.50% |
| 0.09 | 15% | 15% |

**5.2.3 Image with changed contrast (Histogram Equalised):**

Table 7

| Transform | Haar | Co-occurrence |
|---|---|---|
| **Accuracy** | 15% | 40% |

**5.2.4 Rotated Images :**

Table 8

| Rotation | Haar | Co-occurrence |
|---|---|---|
| 2 degree | 73.30% | 2.50% |
| 4 degree | 73.33% | 2.50% |
| 30 degree | 53.33% | 2.50% |

As seen from the table of rotation for the Haar wavelet and Co-occurrence Matrix it can be concluded that the accuracy of co-occurrence matrix method is greatly reduced. It is as less as 2.5% whereas for Haar wavelet it is around 66.66 % (average).

The main reason for this is because Co occurrence matrix applied captures the spatial dependence of wavelet and detail coefficients depending on different directions and distance specified which means that the matrix is totally dependent on the values of the neighboring pixels. When the image is rotated the pixel values are also changed which results in the inaccurate classification results.

## 6. CONCLUSION:

As initiated in the abstract and consolidated progressively in subsequent points, the strength and capabilities of some of the important mathematical descriptions of texture property of an image to produce perceptual quality texture classification are reviewed and compared.

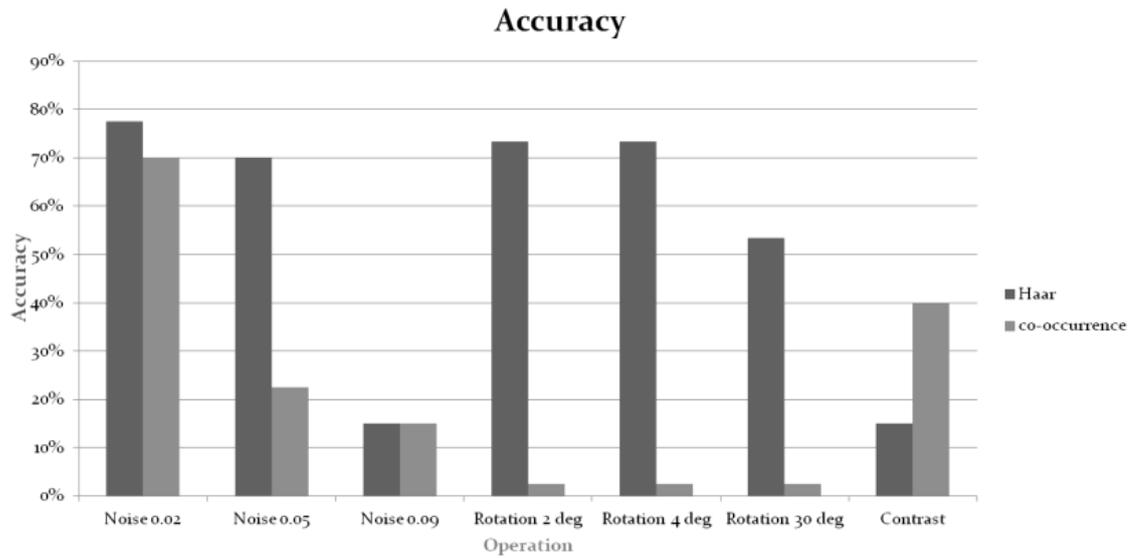

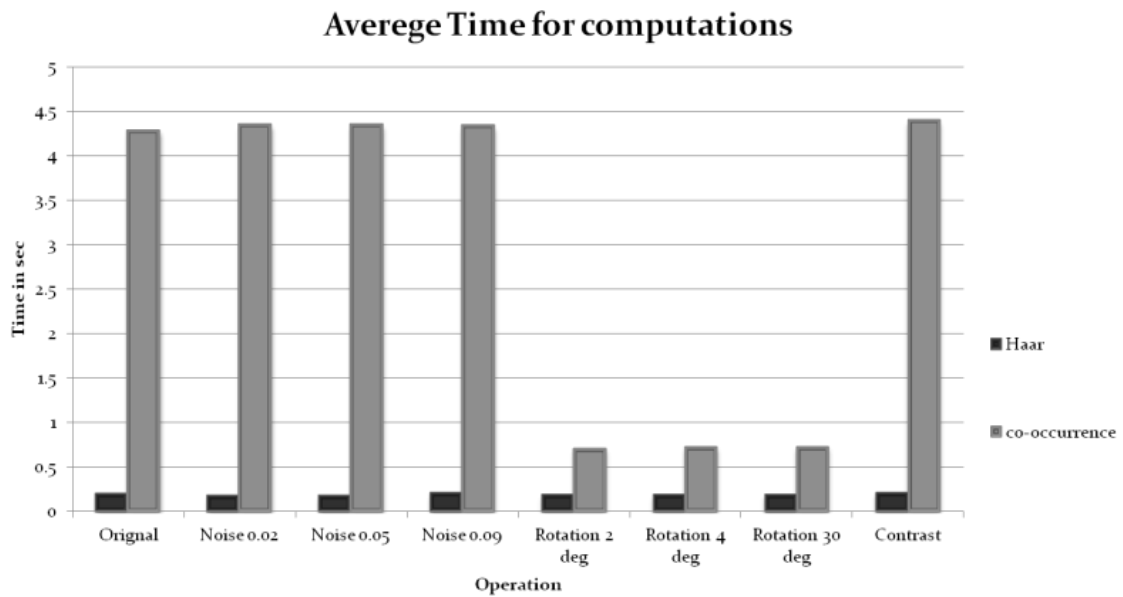

## ACKNOWLEDGEMENTS

We would like to extend our heartfelt gratitude to our Supervisor Mr M. M. Kulkarni for being so helpful and for providing us with his invaluable time and guidance. We would also like to thank Prof. Pooja Kulkarni and Prof. Tornekar without who we would not have been able to successfully complete our project.

## REFERENCES

[1]     Wavelet tutorial by Robi Polikar .

## AUTHORS

Pooja Maknikar

Author graduated as B. Tech Electronics with Digital Signal processing as majors from Vishwakarma Institutes of Technology, Pune. Her final year project was an assistantship under a professor pursuing PhD in Digital Signal Processing. Her professional career outside of academia is working as Project Engineer with WIPRO technologies, Bangalore. Outside of Professional interests, she reads widely and enjoys Trekking.

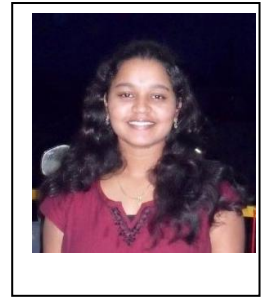

Ashwini Dange

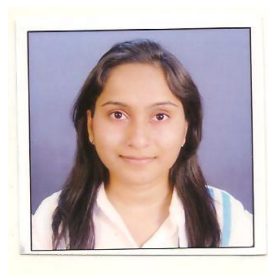

Author is now working as Security Product Developer at Cyberonyx Technologies, Pune. She has completed her B.Tech Electronics from Vishwakarma Institute of Technology, Pune. Besides profession she likes interacting with people and reading books.

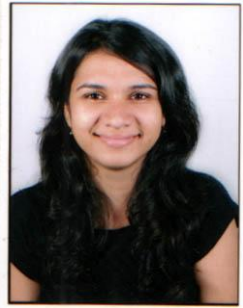

Mugdha Khade

Author has completed her graduation in Electronics and Telecommunication from VIT, Pune. She will be working in Deloitte Consultancy as a Business Technical Analyst. She is currently interning with 'Teach For India' which is a non-profit organisation. Her hobbies include Basketball, reading and travelling

Payal Kulkarni

Author graduated as B. Tech Electronics with Communication as majors from Vishwakarma Institutes of Technology, Pune. Her final year project was an assistantship under a professor pursuing PhD in Digital Signal Processing. Her professional career outside of acadamics is working as a Security Project Developer with Cyberonyx Technologies, Pune. Outside of professional interests she enjoy playing badminton and trevelling.

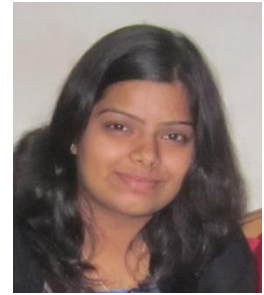